\pdfoutput=1

\documentclass[11pt]{article}

\usepackage[]{acl}

\usepackage{times}
\usepackage{latexsym}

\usepackage[T1]{fontenc}
\usepackage{makecell}
\usepackage{amsmath}
\usepackage{graphicx}
\usepackage{epstopdf} 
\usepackage{multirow}
\usepackage{amssymb}
\usepackage[utf8]{inputenc}

\usepackage{microtype}

\usepackage{inconsolata}

\usepackage{graphicx}

\title{Efficient and Effective Internal Memory Retrieval for LLM-Based Healthcare Prediction}

\author{
Mingchen Li\textsuperscript{\normalfont 2},
Jiatan Huang\textsuperscript{\normalfont 1},
Zonghai Yao\textsuperscript{\normalfont 2},
Hong Yu\textsuperscript{\normalfont 2,3,4}\\
\textsuperscript{1}University of Connecticut \\
\textsuperscript{2}University of Massachusetts, Amherst \\
\textsuperscript{3}University of Massachusetts, Lowell \\
\textsuperscript{4}UMass Chan Medical School
}

\begin{document}
\maketitle
\begin{abstract}

Large language models (LLMs) hold significant promise for healthcare, yet their reliability in high-stakes clinical settings is often compromised by hallucinations and a lack of granular medical context. While Retrieval-Augmented Generation (RAG) can mitigate these issues, standard supervised pipelines require computationally intensive searches over massive external knowledge bases, leading to high latency that is impractical for time-sensitive care. To address this, we introduce Keys-to-Knowledge (K2K), a novel framework that replaces external retrieval with internal, key-based knowledge access. By encoding essential clinical information directly into the model’s parameter space, K2K enables rapid retrieval from internal key–value memory without inference-time overhead. We further enhance retrieval quality through activation-guided probe construction and cross-attention reranking. Experimental results demonstrate that K2K achieves state-of-the-art performance across four benchmark healthcare outcome prediction datasets.
\footnote{The code is available here: \url{https://anonymous.4open.science/r/K2K-2390/README.md}}

\end{abstract}




\section{Introduction}
Large language models (LLMs) have demonstrated significant potential across diverse healthcare applications~\cite{li2022semantic,achiam2023gpt,li2024zero,guo2025deepseek,li2024cancerllm,li2024benchmarking}. However, their deployment in high-stakes clinical environments is often hindered by hallucinations and an inherent difficulty in accessing granular, patient-specific context. While Retrieval-Augmented Generation (RAG) has emerged as a primary strategy to ground these models in external clinical knowledge~\cite{lewis2020retrieval,li2025biomedrag}, existing approaches—which typically retrieve from structured knowledge graphs~\cite{li2023understand,zhang2025survey}, unstructured documents~\cite{jin2025search}, or self-generated knowledge- introduce a significant computational burden.
\begin{figure}[t]
        \centering
        \includegraphics[width=1\columnwidth]{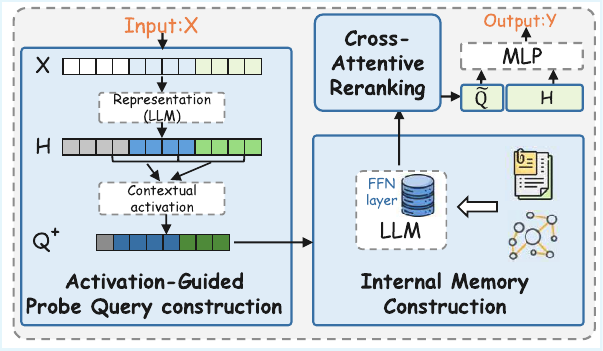}
	\caption{ Overview of K2K. 
    The input X consists of longitudinal EHR diagnostic codes, and the output  y represents healthcare prediction results. K2K first constructs an internal memory within the LLM by injecting external medical knowledge through the Internal Memory Construction module. Given the input, K2K then employs Activation-Guided Probe Construction to build a probe query that retrieves relevant information from the internal memory. Finally, the Cross-Attentive Reranking module dynamically integrates diverse retrieved knowledge. The aggregated memory knowledge, together with the learned input embeddings, is used for the final healthcare prediction.
    }
	\label{con:K2K_figure1}
\end{figure}

Recent work~\cite{su2025parametric} identifies two critical bottlenecks in these traditional pipelines. First, injecting external knowledge via input prompts expands context length, which escalates inference costs and limits scalability. Second, constructing high-quality retrievers remains a major hurdle; supervised retrieval requires extensive labeled query–context pairs, while structured retrieval often relies on costly graph searches or oversimplified heuristics~\cite{li2024zero} that sacrifice accuracy for coverage. These challenges create substantial overhead in data annotation and computation, particularly when navigating large-scale, heterogeneous medical knowledge bases. Ultimately, the requirement for exhaustive external searches results in high end-to-end latency, a prohibitive flaw in time-sensitive clinical settings where rapid decision-making is vital.

To overcome these limitations, we introduce Keys-to-Knowledge (K2K), a novel framework that bypasses external retrieval by facilitating internal, key-based knowledge access directly within the model’s architecture. As prior work has demonstrated~\cite{geva2020transformer}, the keys within the feed-forward network (FFN) layers of transformer-based models implicitly store factual knowledge. A promising research direction is to retrieve these query-relevant FFN keys as internal knowledge, enabling direct access to information without the burden of long external contexts or the complexity of structured knowledge base searches.

However, using the raw query alone to retrieve top-$k$ keys, without incorporating contextual activation signals, does not guarantee accuracy or relevance. Our preliminary experiments indicate that disparate queries often yield highly similar retrieved keys, suggesting that the resulting probe representations lack discriminative power. Specifically, these representations tend to obscure vital semantic distinctions, a phenomenon also observed in recent studies~\cite{xiao2025activation}, which ultimately degrades retrieval effectiveness. Moreover, internal key-space retrieval faces two significant hurdles: \textit{Lack of Grounding}: The retrieved key vectors are latent and ungrounded, lacking the explicit provenance found in documents or knowledge graphs. \textit{Static Retrieval}: The process remains non-adaptive, lacking explicit semantic signals to guide the dynamic reweighting of retrieved knowledge for specific downstream tasks.

To address these challenges, K2K retrieves key-based knowledge from LLM purposefully infused with external clinical information through three core modules. As illustrated in Figure~1, \textit{Internal Memory Construction} grounds latent keys with external clinical knowledge, \textit{Activation-Guided Probe Construction} enhances probe discriminability by incorporating contextual activation signals, and \textit{Cross-Attentive Reranking} enables adaptive, task-aware reweighting of retrieved knowledge.

Our contributions are summarized as follows:

\textbf{Internal Memory Construction}: We transform the pre-trained language model into a retrieval memory by leveraging the keys stored in FFN layers. For domain-specific knowledge absent from the pre-training corpus, we employ LoRA~\cite{hu2021lora} to inject new information into the parameter space. This approach mitigates reliance on external retrievers and eliminates the latency overhead of long input contexts.

\textbf{Activation-Guided Probe Construction}: To ensure accurate retrieval from internal memory, we design a probe-query mechanism that identifies critical tokens and scarce outlier features during inference. We utilize an activation bias, computed via a diagonal approximation of the Mahalanobis distance, to emphasize query vectors with high discriminative power, thereby balancing per-dimension variance and improving retrieval precision.

\textbf{Cross-Attentive Reranking}: To account for varying relevance and structural dependencies across retrieved knowledge, we introduce a cross-attentive reranking mechanism. This component dynamically integrates and reweights multi-source internal knowledge conditioned on the specific clinical query, ensuring context-dependent integration for downstream tasks.

\section{Preliminaries}
\begin{figure*}[t]
        \centering
        \includegraphics[width=1.9\columnwidth]{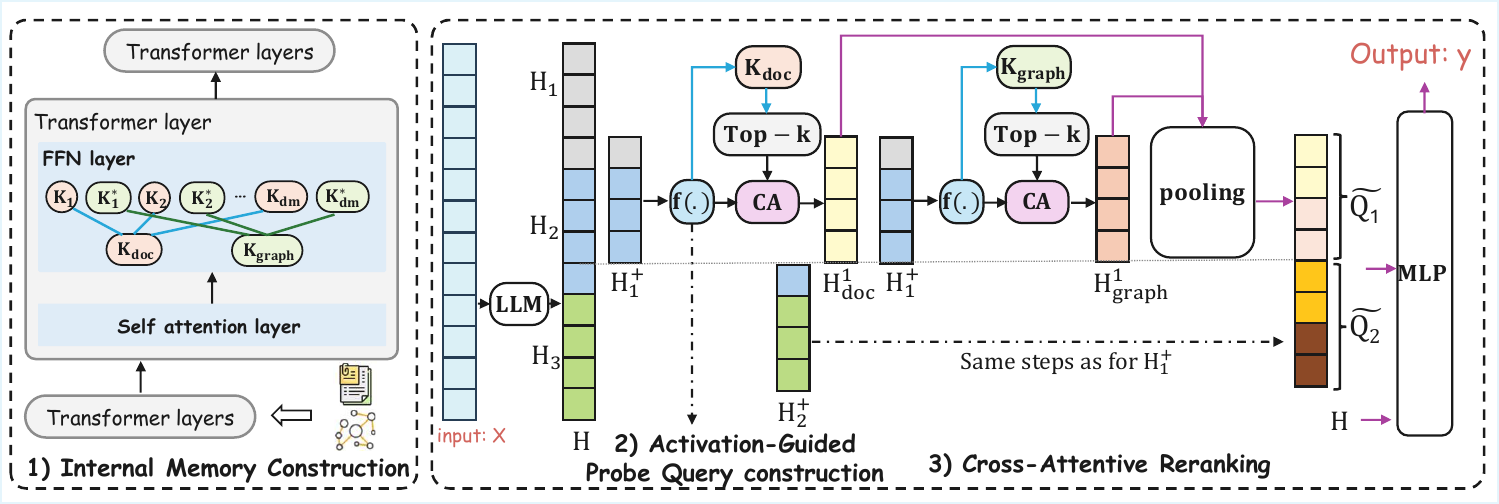}
	\caption{Overview of the K2K framework, consisting of three steps: (1) Retrieval Memory Construction builds $K_{\text{doc}} = [K_1, K_2, \cdots, K_{dm}]$ and $K_{\text{graph}} = [K_1^*, K_2^*, \cdots, K_{dm}^*]$; (2) Activation-Guided Probe Query Construction  (Blue box function f(.)) enhances the query representation for key retrieval from  $K_{\text{doc}}$ and $K_{\text{graph}}$; (3) Cross-Attentive Reranking retrieves relevant document knowledge $H_{\text{doc}}^w$ and graph knowledge $H_{\text{graph}}^w$ for the enhanced query $Q_w^+ = f(H_w^+)$, and integrates them with the original representation $H$ for final prediction. Here, $t \in {1, 2, 3}$.}
	\label{con:K2K_framework}
\end{figure*}

\textbf{FFN Architecture}: In transformer-based architectures~\cite{vaswani2017attention}, the feed-forward network (FFN) operates in tandem with the self-attention mechanism to transform hidden representations. Each FFN layer acts as a position-wise function, processing each input vector independently. Given an input vector $\mathbf{x} \in \mathbb{R}^d$ typically sourced from a preceding attention layer, the output of the FFN, denoted as $\mathrm{FF}(\mathbf{x})$, is defined as:

\begin{equation}\mathrm{FF}(\mathbf{x}) = \sigma(\mathbf{x} W_1 + \mathbf{b}_1) W_2 + \mathbf{b}_2 \label{ffnw}\end{equation}

where $W_1 \in \mathbb{R}^{d \times d_{ff}}$ and $W_2 \in \mathbb{R}^{d_{ff} \times d}$ are weight matrices, $\mathbf{b}_1$ and $\mathbf{b}_2$ are bias terms, and $\sigma(\cdot)$ denotes a non-linear activation function (e.g., ReLU or GeLU).

\textbf{Key-Value Memory Interpretation}: Following Geva et al.~\cite{geva2020transformer}, the FFN can be viewed as an associative key-value memory system. In this view, the weight matrices represent the stored knowledge: $W_1$ serves as the keys and $W_2$ as the values. Specifically, the FFN operation can be reformulated as:

\begin{equation}\mathrm{FF}(\mathbf{x}) = \sum_{i=1}^{d_{ff}} \sigma(\mathbf{x} \cdot \mathbf{k}_i) \mathbf{v}_i\end{equation}

where $\mathbf{k}_i$ (the $i$-th column of $W_1$) is a key vector that captures specific semantic patterns, and $\mathbf{v}_i$ (the $i$-th row of $W_2$) is the corresponding value vector. This interpretation suggests that a model's factual knowledge is stored within these parameter-based keys, providing the theoretical foundation for K2K to treat the model's internal parameter space as a retrievable knowledge base.

\textbf{Knowledge Infusion via LoRA}: To incorporate domain-specific medical knowledge not present in the pre-training corpus, we employ Low-Rank Adaptation (LoRA)~\cite{hu2021lora}. LoRA represents weight updates as the product of two low-rank matrices, $A$ and $B$. Under this formulation, the adapted FFN becomes:

\begin{equation}\mathrm{FF}(\mathbf{x}) = \sigma(\mathbf{x}(W_1 + A_1 B_1) + \mathbf{b}_1)(W_2 + A_2 B_2) + \mathbf{b}_2 \label{ffn_lora}\end{equation}

In the context of K2K, the effective internal memory consists of the updated key and value matrices, defined as $K^\top = W_1 + A_1 B_1$ and $V = W_2 + A_2 B_2$, respectively. By freezing the pre-trained weights and only updating the low-rank adapters, we efficiently infuse clinical knowledge into the model's internal memory space while maintaining its general reasoning capabilities.

\section{Methodology}

As illustrated in Figure~\ref{con:K2K_framework}, the K2K framework operates in three distinct stages: (1) \textbf{Retrieval Memory Construction}, where domain-specific knowledge is infused into the model’s internal parameter space; (2) \textbf{Activation-Guided Probe Construction}, which identifies salient context features for precise knowledge matching; and (3) \textbf{Cross-Attentive Reranking}, which dynamically integrates the retrieved internal knowledge for the final prediction.

\subsection{Retrieval Memory Construction}
The retrieval memory in K2K encompasses two primary types of clinical information: 

\textbf{Document-Level Memory}: We utilize a pre-trained large language model ($\mathcal{M}_{\text{base}}$) as our foundational backbone. Rather than performing computationally expensive continued pre-training on the entire model, we leverage a domain-adapted model ($\mathcal{M}_{\text{domain}}^{\text{doc}}$). Following the memory interpretation of FFNs, we treat the Keys ($W_1$) within the $l$-th Transformer layer of $\mathcal{M}_{\text{domain}}^{\text{doc}}$ as the internal representation of document-level knowledge, denoted as $K_{\text{doc}}^l$.

\textbf{Graph-Level Memory}: To incorporate structured knowledge, we linearize triples from a medical knowledge graph into textual descriptions (e.g., "The relationship between [head] and [tail] is [relation]"). We then apply LoRA-based adaptation to $\mathcal{M}_{\text{domain}}^{\text{doc}}$ using this organized triples dataset. The resulting LoRA adapter matrices $A_1 B_1$ (per Equation~\ref{ffn_lora}) from the FFN layers are designated as the structured knowledge source for layer $l$, denoted as $K_{\text{graph}}^l$.

\subsection{Activation-Guided Probe Query Construction}
\label{con:M-Probe}
As noted by \cite{xiao2025activation}, standard probe queries often rely on mean pooling, which disperses attention across all tokens and fails to capture core semantics. To address this, we propose a \textbf{Contextual Activation Weight} to identify the most informative query vectors within a context window.

Given a sequence of query vectors\footnote{The query vectors for window 
$w$ are the embeddings encoded by the LLM from the input sequence 
$X$.} $H_w = [h_1^w, h_2^w, \dots, h_L^w]$ for window $w$ with length $L$ in the input  $X$, we first compute the statistical mean $\bar{z}^w$:

\begin{equation}\bar{z}^w = \frac{1}{L} \sum_{j=1}^{L} h_j^w\end{equation}

While prior work utilized Euclidean distance to weight tokens, that approach assumes uniform importance across all embedding dimensions. To better account for per-dimension variance and increase sensitivity to deviations in low-variance directions, we employ a diagonal approximation of the Mahalanobis distance:

\begin{equation}
\phi_j^w \approx \sqrt{ \sum_{d=1}^{D} \frac{(h_{j,d}^w - \bar{z}_d^w)^2}{\sigma_d^2} }
\end{equation}

where $\sigma_d^2$ denotes the variance computed across all tokens within the window along the $d$-th embedding dimension, $\bar z_d^w$ denotes the $d$-th component of the mean vector $\bar z^w$.
We then normalize these scores to obtain a soft attention distribution $\alpha_j^w$:

\begin{equation}
\alpha_j^w = \frac{\phi_j^w}{\sum_{j=1}^{L} \phi_j^w}
\end{equation}

Finally, the enhanced probe vector $Q_w$ is constructed by aggregating token vectors $h_j^w$ using these weights, effectively emphasizing semantically grounded "anchor" tokens: then normalize these

\begin{equation}Q_w = f(H_w) = \sum_{j=1}^{L}\alpha_j^w \cdot h_j^w\end{equation}

\subsection{Cross Attention Reranking}
\label{con:cross_windown}
To perform cross-attention reranking, following RETRO~\cite{borgeaud2022improving}, we first split the representation $H$ of input sentence into a sequence of $w{-}1$ windows, denoted as $\{H_1^+, H_2^+, \dots, H_{w-1}^+\}$. $H^+_w$ represents the  query embeddings constructed by concatenating the last token of window $w$ and the first $L-1$ tokens of window $w+1$.
For each window $w$, we generate an enhanced query representation $Q_w^+ = f(H_w^+)$. We then retrieve the top-$k$ most relevant vectors from the internal document and graph memories based on similarity scores:

\begin{equation}
\small
\begin{array}{c}
K_{\text{doc}}^w = \text{top-}k\left( \text{sim}(Q_w^+, K_{\text{doc}}^l) \right) \\
K_{\text{graph}}^w = \text{top-}k\left( \text{sim}(Q_w^+, K_{\text{graph}}^l) \right)
\end{array}
\label{con:topk}
\end{equation}

We apply a Cross-Attention (CA) mechanism to rerank and refine these vectors, yielding the final retrieved document knowledge $H_{\text{doc}}^{w}$ and graph knowledge $H_{\text{graph}}^{w}$:

\begin{equation}
\small
\begin{array}{c}
{H}_{\text{doc}}^{w} = \text{CA}(Q_w^+, K_{\text{doc}}^{\text{w}}, V_{\text{doc}}^{\text{w}}) \\
{H}_{\text{graph}}^{w}  = \text{CA}(Q_w^+, K_{\text{graph}}^{\text{w}}, V_{\text{graph}}^{\text{w}})
\end{array}
\end{equation}

The retrieved vectors are normalized via a pooling function $P(\cdot)$ and fused via concatenation:

\begin{equation}
\tilde{Q}_w =[ P({H}_{\text{doc}}^{w}) ; P({H}_{\text{graph}}^{w})]
\end{equation}
We then aggregate all window-level fused representations together with the input sentence representation $H$ and feed the combined representation into an MLP for final prediction. The loss is defined as:
\begin{equation}
\small
\mathcal{L}_{\text{cls}} = \text{CrossEntropy}\left( \text{MLP}([H;\;\tilde{Q}_1 \;  \dots \;\tilde{Q}_{w-1} ]),\; y \right)
\end{equation}
where $y$ denotes the ground truth label.

\section{Experiments}

\subsection{Problem Formulation}

We define the clinical history of a patient as a sequence of hospital visits $V = \{v_1, v_2, \dots, v_{|V|}\}$. Each visit $v_i$ is associated with a set of International Classification of Diseases (ICD) codes $C_i = \{c_{i,1}, c_{i,2}, \dots \}$, where each code $c \in C_i$ represents a specific diagnosis or procedure. For each code, we utilize its corresponding clinical descriptor $s$, provided as a short text snippet (e.g., "Acute myocardial infarction").The goal of our model is to predict a binary clinical outcome $y \in \{0, 1\}$ based on the historical context. We evaluate our framework on two critical tasks:

\textbf{Mortality Prediction}: $y_i$ indicates whether the patient's death occurs during the subsequent visit $v_{i+1}$.

\textbf{Readmission Prediction}: Following the protocol in KARE~\cite{jiang2024reasoning}, $y_i$ predicts whether the patient will be readmitted to the hospital within $\alpha = 15$ days.


\subsection{Dataset}

We evaluate K2K on the publicly available MIMIC-III~\cite{johnson2016mimic} and MIMIC-IV~\cite{johnson2020mimic} datasets. The statistics for the processed data are summarized in Table~\ref{con:mimic data}.

\begin{table}[ht]
	\centering
	
	\renewcommand\arraystretch{1.3}
	\scalebox{0.9}{
	\begin{tabular} {ccccc}
		\hline 
		
		\hline	
		&   III-Mort   & III-Read   & IV-Mort   & IV-Read    \\ 
		\hline		
		Train &  7,777   &  7,777 & 100,125  &  10,0125  \\ 
		Test&    953  &    953  &12,667   &   12,667  \\ 
        	Dev &  978   &    978   & 12,547 &  12,547  \\ 
		\hline
	\end{tabular}}
		\caption{ Datasets Statistics, Mort refers to the Mortality. III refers to the MIMIC-III. Read refers to readmission. }
	\label{con:mimic data}
\end{table}

We employ a 0.8/0.1/0.1 split ratio for training, validation, and testing, respectively. To prevent data leakage, the split is grouped by patient ID, ensuring that all visits from a single patient are confined to a single subset. While prior work such as KARE~\cite{jiang2024reasoning} utilizes a random subset of MIMIC-IV, we utilize the full dataset to better simulate the scale and heterogeneity of real-world clinical settings.

	
		


\subsection{Baselines and Evaluation Metrics}

We compare K2K against a diverse range of competitive baselines:

\textbf{Sequential Models}: GRU~\cite{chung2014empirical}, RETAIN~\cite{choi2016retain}, Deepr~\cite{nguyen2016deepr}, AdaCare~\cite{ma2020adacare}, StageNet~\cite{gao2020stagenet}, and TCN~\cite{bai2018empirical}.

\textbf{Retrieval-Based Models}: KARE~\cite{jiang2025reasoning}, the current state-of-the-art for healthcare prediction, and Standard RAG~\cite{li2024benchmarking}, which uses Contriver~\cite{izacard2021unsupervised} to retrieve relevant patient examples.

\textbf{Generative Knowledge Models}: Prompt-Based Retrieval~\cite{frisoni2024generate}, which leverages in-context learning to instruct an LLM to generate task-relevant medical knowledge.

\textbf{Evaluation Metrics}: Following standard practices in clinical predictive modeling~\cite{jiang2025reasoning, jiang2023graphcare}, we report the F1-score, Jaccard Similarity, AUPRC, and AUROC. Detailed implementation parameters and hardware configurations are provided in Appendix~\ref{con:Implementation Detail}.



\subsection{Main Results}

Table~\ref{con:main_results} presents the main results and highlights several key observations for two LLMs, BioMistral-7B and Meditron3-Qwen2.5-7B: (1) K2K demonstrates consistently strong performance across datasets and tasks, achieving the highest average scores among the evaluated methods. (2)  Baseline retrieval methods fail to capture the semantic nuances of the input. Although KARE enhances retrieval by combining relevant documents with the shortest paths from the graph, such paths may overlook critical relational information. In contrast, our method retrieves key knowledge directly from the language model’s internal knowledge store, enabling more comprehensive and context-aware retrieval. (3) We find that BioMistral-7B performs worse than traditional machine learning models when the input contains discontinuous or complex diagnoses and suffers from class imbalance between positive and negative samples. This is also observed by \citet{gao2025uncertainty}.  While Meditron3-Qwen2.5-7B exhibits stronger performance. By introducing document-level knowledge and graph-based knowledge into the language model, our method achieves improved performance. For example, K2K outperforms LLMs without retrieval mechanisms on the mortality prediction task using the MIMIC-IV dataset.  (4) Under BioMistral-7B, we found that prompt-based retrieval outperforms standard RAG by retrieving knowledge from external documents, enabling the language model to generate more useful information that improves the classification results,  as evidenced by improvements in AUPRC and AUROC on the Mortality-MIMIC-III dataset. Additional efficiency results are reported in Appendix \ref{con:Retrieval_Inference_Efficiency}, showing that K2K achieves competitive average performance while being substantially more efficient in retrieval time compared with KARE and prompt-based methods on the test set of MIMIC-III.

\begin{table*}[t]
\centering
\setlength{\tabcolsep}{4.5pt}
\renewcommand\arraystretch{1.08}
\scalebox{0.7}{
\begin{tabular}{c c ccccc ccccc}
\Xhline{0.5pt}

& &
\multicolumn{5}{c}{\textbf{Mortality-MIMIC-III}} &
\multicolumn{5}{c}{\textbf{Readmission-MIMIC-III}} \\

\textbf{Type} & \textbf{Model}
& \textbf{F1} & \textbf{Jac.} & \textbf{AUPRC} & \textbf{AUROC} & \textbf{Avg}
& \textbf{F1} & \textbf{Jac.} & \textbf{AUPRC} & \textbf{AUROC} & \textbf{Avg} \\
\hline

\multirow{6}{*}{\scriptsize ML}
& GRU~\cite{chung2014empirical}
& 13.87 & 7.45 & 8.03 & 53.50 & 20.71
& 68.28 & 51.84 & 52.94 & 50.00 & 55.77 \\

& RETAIN~\cite{choi2016retain}
& 13.73 & 7.37 & 9.57 & 54.86 & 21.38
& 45.88 & 23.48 & 54.11 & 51.29 & 43.69 \\

& Deepr~\cite{nguyen2016deepr}
& 13.87 & 7.45 & 7.58 & 51.66 & 20.14
& 68.28 & 51.84 & 51.68 & 49.70 & 55.38 \\

& AdaCare~\cite{ma2020adacare}
& 12.90 & 6.89 & 7.80 & 50.69 & 19.57
& 63.49 & 46.51 & 52.83 & 52.27 & 53.77 \\

& StageNet~\cite{gao2020stagenet}
& 9.97 & 5.25 & 7.10 & 47.14 & 17.37
& 51.56 & 34.74 & 50.38 & 48.27 & 46.24 \\

& TCN~\cite{bai2018empirical}
& 11.28 & 5.97 & 6.76 & 45.81 & 17.45
& 65.46 & 48.66 & 49.84 & 47.65 & 52.90 \\

\Xhline{0.4pt}

\multirow{12}{*}{\scriptsize LLM}
& \multicolumn{11}{l}{\footnotesize Fine-tuned (LLM: BioMistral-7B)} \\
\cline{2-12}

& w/o retriever
& 16.00 & 8.69 & 11.61 & 59.40 & 23.92
& 69.17 & 52.87 & 59.07 & 54.61 & 58.93 \\

& KARE~\cite{jiang2025reasoning}
& 18.01 & 9.90 & 9.72 & 56.65 & 23.57
& 61.64 & 44.55 & 56.67 & 50.97 & 53.46 \\

& Standard RAG~\cite{li2024benchmarking}
& 11.94 & 6.34 & 9.34 & 54.19 & 20.45
& 69.73 & 53.52 & 57.09 & 52.99 & 58.33 \\

& Prompt-based~\cite{frisoni2024generate}
& 15.05 & 8.13 & 10.78 & 58.72 & 23.17
& 66.51 & 49.82 & 54.19 & 49.71 & 55.06 \\

& \textbf{K2K (Ours)}
& 18.55 & 10.22 & 15.22&61.05 & \textbf{26.26}
& 69.31 & 53.03 & 62.49 & 56.64 &  \textbf{60.37} \\
& GRPO
& 16.90 & 9.10 & 13.80 & 58.40 & 24.55
& 66.80 & 51.20 & 58.90 & 54.70 & 57.90 \\
& GRPO——advance
& 18.91 & 10.05 & 15.74 & 61.54 & 26.56
& 69.12 & 53.97 & 63.18 & 57.20 & 60.87 \\
\cline{2-12}
& \multicolumn{11}{l}{\footnotesize Fine-tuned (LLM: Meditron3-Qwen2.5-7B)} \\
\cline{2-12}

& w/o retriever
& 13.78 & 7.40 & 7.00 & 48.90 & 19.27
& 68.18 & 51.72 & 57.72 & 53.01 & 57.66 \\

& KARE~\cite{jiang2025reasoning}
& 11.67 & 6.19 & 10.38 & 56.19 & 21.11
& 59.45 & 42.30 & 55.66 & 50.89 & 52.08 \\
&Standard RAG~\cite{li2024benchmarking} &16.35  & 8.90  &8.43      & 54.11  &21.95   & 63.00& 45.98 & 58.14   &    53.90 &  55.26    \\ 
& Prompt-based~\cite{frisoni2024generate}
& 15.44 & 8.36 & 12.01 & 54.29 & 22.52
& 60.15 & 43.01 & 60.25 & 55.38 & 54.70 \\

& \textbf{K2K (Ours)}
& 17.27 & 9.45 & 10.29 & 54.56 & \textbf{22.89}
& 68.78 & 52.42 & 59.88 & 56.08 & \textbf{59.29} \\

\Xhline{0.5pt}

& &
\multicolumn{5}{c}{\textbf{Mortality-MIMIC-IV}} &
\multicolumn{5}{c}{\textbf{Readmission-MIMIC-IV}} \\

\textbf{Type} & \textbf{Model}
& \textbf{F1} & \textbf{Jac.} & \textbf{AUPRC} & \textbf{AUROC} & \textbf{Avg}
& \textbf{F1} & \textbf{Jac.} & \textbf{AUPRC} & \textbf{AUROC} & \textbf{Avg} \\
\hline

\multirow{6}{*}{\scriptsize ML}
& GRU~\cite{chung2014empirical}
& 3.20 & 1.62 & 1.66 & 53.71 & 15.05
& 59.28 & 42.13 & 57.38 & 56.58 & 53.84 \\

& RETAIN~\cite{choi2016retain}
& 2.78 & 1.41 & 1.43 & 47.18 & 13.20
& 66.77 & 50.12 & 51.44 & 49.61 & 54.48 \\

& Deepr~\cite{nguyen2016deepr}
& 2.86 & 1.46 & 1.57 & 51.48 & 14.34
& 68.13& 51.66 & 52.27 & 50.44 & 55.62 \\

& AdaCare~\cite{ma2020adacare}
& 2.98 & 1.52 & 1.53 & 51.41 & 14.36
& 47.96 & 31.54 & 52.12 & 50.38 & 45.50 \\

& StageNet~\cite{gao2020stagenet}
& 2.96 & 1.50 & 1.60 & 51.11 & 14.29
& 48.11 & 31.67 & 50.74 & 48.67 & 44.80 \\

& TCN~\cite{bai2018empirical}
& 2.92 & 1.48 & 1.63 & 54.17 & 15.05
& 53.32 & 36.35 & 51.33 & 49.62 & 47.66 \\

\Xhline{0.4pt}

\multirow{12}{*}{\scriptsize LLM}
& \multicolumn{11}{l}{\footnotesize Fine-tuned (LLM: BioMistral-7B)} \\
\cline{2-12}
& w/o retriever
& 1.08 & 0.50 & 1.30 & 44.61 & 11.87
& 61.30 & 44.20 & 67.86 & 65.83 & 59.80 \\

& KARE~\cite{jiang2025reasoning}
& 1.33 & 0.67 & 1.46 & 49.55 & 13.25
& 61.75 & 44.67 & 67.09 & 65.44 & 59.74 \\

& Standard RAG~\cite{li2024benchmarking}
& 2.45 & 1.61 & 2.74 & 55.92 & 15.68
& 60.95 & 43.84 & 68.51 & 66.64 & 59.98 \\

& Prompt-based~\cite{frisoni2024generate}
& 3.16 & 1.60 & 1.49 & 48.26 & 13.63
& 61.02 & 43.91 & 68.89 & 67.02 & 60.21 \\

& \textbf{K2K (Ours)}
&  6.61 & 3.42 & 2.93 & 66.50 & \textbf{19.87}
& 63.75 & 46.79& 68.67 & 66.47 & \textbf{61.42} \\
\cline{2-12}
& \multicolumn{11}{l}{\footnotesize Fine-tuned (LLM: Meditron3-Qwen2.5-7B)} \\
\cline{2-12}
& w/o retriever
&3.41  &1.73 &  2.72&  63.52 &  17.85
& 56.92  & 39.79 & 67.76  & 65.86   &57.58\\

& KARE~\cite{jiang2025reasoning}
&4.41  & 2.25&  1.96& 59.20  & 16.96 
&  64.62 & 47.74 & 68.50  & 66.76 & 61.91 \\

& Standard RAG~\cite{li2024benchmarking}
& 4.47 &2.29 & 2.35 &64.87   & 18.49 
&   65.09& 48.25 & 67.68  &65.78  & 61.70 \\

& Prompt-based~\cite{frisoni2024generate}
& 4.17 & 2.13&  2.17&  59.68 &  17.04
&  57.71 & 40.55 &  66.91 & 64.84 &  57.50\\
& \textbf{K2K (Ours)}
& 5.10 & 2.61& 2.59 & 63.85  & \textbf{ 18.54}
& 67.11  & 50.50 & 67.60  & 66.28 &\textbf{ 62.87 }\\
\Xhline{0.5pt}
\end{tabular}
}

\caption{
Comparative analysis of mortality and readmission prediction on MIMIC-III and MIMIC-IV.
\textbf{Avg} denotes the arithmetic mean of F1, Jaccard, AUPRC, and AUROC for each model on the corresponding dataset. Boldface indicates the highest average score under each LLM.
}
\label{con:main_results}
\end{table*}

\section{Analysis}
To further evaluate the effectiveness of our framework, we conduct a series of analyses based on different components of our model. First, we investigate the impact of different knowledge sources by introducing two ablations: K2K without document knowledge and K2K without graph knowledge (Section~\ref{con:impact_knowledge_source}). We also assess the performance of directly using the LLM with its internal knowledge to make predictions, in order to validate the effectiveness of our key knowledge retrieval framework, which leverages cross-window attention (Section~\ref{con:direct_un_direct_knowledge_source}).
Next, we compare different query representation strategies to demonstrate the effectiveness of our proposed diagonal approximation of the Mahalanobis distance (Section~\ref{con:query_re_stra}).
Finally, we analyze the effect of retrieving knowledge from different LLM layers (Section~\ref{con:layers}). In this section, we use BioMistral-7B as the base LLM.

For additional experiments on K2K, including the effect of different chunk sizes, the impact of the hyperparameter top-$k$ in Equation~\ref{con:topk}, and analyses of retrieval   efficiency across various retrieval methods and pipelines, please refer to Appendix~\ref{con:chunk_size_compare},\ref{con:topk_ana},\ref{con:Retrieval_Inference_Efficiency}.


\subsection{Impact of Different Knowledge Source}
\label{con:impact_knowledge_source}

\begin{table}[ht]

	\centering
	\renewcommand\arraystretch{1.3}
	\scalebox{0.6}{
	\begin{tabular} {c|c|cccc}
    \Xhline{0.5pt} 
		\hline 
       &Model &F1 &Jaccard& AUPRC &AUROC  \\
       \hline
  
             \multirow{3}{*}{Mortality-III}  & K2K   &  18.55   &  10.22 &  \textbf{15.22  }  &  \textbf{61.05    }   \\
                            &  K2K w/o graph &\textbf{ 20.48} & \textbf{11.40}& 13.18 &  60.54     \\
                   &  K2K w/o document  &  16.66&9.09 & 10.52 &  55.72      \\
        \hline
                 \multirow{3}{*}{Mortality-IV}  & K2K Ours& \textbf{6.61 }  &   \textbf{ 3.42} &   \textbf{2.93}    &  \textbf{66.50}  \\
                                    &   K2K w/o graph & 4.50 &2.30 & 2.51 &60.86        \\
                                    &  K2K w/o document  &3.57& 1.82& 2.71 &66.41  \\
          \hline
             \multirow{3}{*}{Readmission-III}  & K2K    & 69.31 &   53.03& \textbf{  62.49  } &      \textbf{   56.64 }          \\
                            &  K2K w/o graph    &  \textbf{ 70.95}  &  \textbf{ 54.98} &  60.87 &  54.55     \\
                   &  K2K w/o document   & 69.74  &53.54  & 61.93  &  56.36       \\
        \hline
    
                 \multirow{3}{*}{Readmission-IV}  & K2K Ours   &    \textbf{ 63.75} &  \textbf{46.79 } &   \textbf{68.67 } &   \textbf{66.47     }     \\
                                    &   K2K w/o graph      &55.31   & 38.23 &  66.14 & 64.06       \\
                   &  K2K w/o document     &  56.95 &39.81  & 55.43  & 64.68       \\
                  \hline
                   \Xhline{0.5pt} 

	\end{tabular}
 }
		\caption{Results of different knowledge sources in K2K}
	\label{con:knowledge_source}
\end{table}

Table~\ref{con:knowledge_source} presents the results of K2K using different knowledge sources. Specifically, K2K w/o document refers to the variant of K2K that uses only the retrieved graph knowledge $K_{graph}^t$, as described in Section~\ref{con:cross_windown}. To ensure a fair comparison, the only difference between K2K and its ablated versions (w/o document or w/o graph) is the type of knowledge source used. From Table~\ref{con:knowledge_source}, we observe that the performance of K2K drops when either document or graph knowledge is removed, especially on the MIMIC-III dataset.  Moreover, although K2K w/o graph achieves a higher F1 score, its lower AUPRC and AUROC suggest that it may overfit to a specific threshold and lacks robustness in distinguishing positive cases across varying decision boundaries. In contrast, K2K achieves more balanced performance across all metrics, indicating better generalization and retrieval effectiveness.

\subsection{Direct Use vs. Retrieved Use of Pre-trained Knowledge}
\label{con:direct_un_direct_knowledge_source}

\begin{table}[ht]

	\centering
	\renewcommand\arraystretch{1.3}
	\scalebox{0.6}{
	\begin{tabular} {c|c|cccc}
    \Xhline{0.5pt} 
		\hline 
       &Model &F1 &Jaccard& AUPRC &AUROC  \\
       \hline
  
             \multirow{5}{*}{Mortality-III}  & K2K   &  \textbf{18.55}   &\textbf{ 10.22 } &  \textbf{15.22  }& \textbf{61.05}   \\
                        &   LLM  &  4.49  & 2.29 & 8.67   &  55.62    \\
                        & LLM+Doc   &16.00     &  8.69&  11.61  &    59.40  \\
                    & LLM+Graph  & 4.50    &2.29   & 8.67   &   55.62   \\
                    & LLM+Doc+Graph  & 16.00   &  8.70& 11.61   &   59.41   \\
        \hline
         \multirow{5}{*}{Readmission-III}  & K2K   &    69.31 & 53.03  & \textbf{62.49  } &  \textbf{56.64}  \\
                        &   LLM &  64.10   & 47.17  & 60.81   &  54.57  \\
                        & LLM+Doc  & 69.17    & 52.87  & 59.07   &54.61    \\
                    & LLM+Graph  &  44.31   & 28.46  & 56.57   & 48.87   \\
                    & LLM+Doc+Graph & \textbf{ 70.81 }  &\textbf{ 54.81}  & 61.51   &  54.70  \\
        \hline
           \multirow{5}{*}{Mortality-IV}  & K2K   & \textbf{ 6.61}  & \textbf{3.42} &  \textbf{2.93}   & \textbf{66.50}    \\
                        &   LLM  &   2.05  & 1.03 & 1.59    &   51.64  \\
                        & LLM+Doc  &  1.08  & 0.50 &   1.30  &  44.61   \\
                    & LLM+Graph &   3.24 & 1.60 & 1.52    &   50.08  \\
                    & LLM+Doc+Graph  & 1.08   & 0.55 &  1.30   &  44.61   \\
                      \hline
                     \multirow{5}{*}{Readmission-IV}  & K2K   & \textbf{63.75   } & \textbf{46.79}  &  \textbf{ 68.67}  &  \textbf{66.47}   \\
                        &   LLM  &    60.06  &42.92   & 66.15    &  64.64   \\
                        & LLM+Doc  & 61.30   &44.20   &   67.86  &  65.83   \\
                    & LLM+Graph &   48.97&32.43  & 50.80    & 48.30    \\
                    & LLM+Doc+Graph &  54.86&   37.80&  51.57  &   49.93   \\
        \hline

	\end{tabular}
 }
		\caption{Comparison of Knowledge-Enhanced Models on Mortality and Readmission Prediction (MIMIC-III/IV). LLM refers to Mixtral-7B. LLM+Doc denotes BioMixtral-7B, which is obtained by further training Mixtral-7B on a medical corpus. LLM+Graph refers to Mixtral-7B adapted to graph-based knowledge using LoRA. LLM+Doc+Graph represents BioMixtral-7B further adapted to graph knowledge via LoRA.}
	\label{con:knowledge_enhanced}
\end{table}

Table~\ref{con:knowledge_enhanced} shows the results of the experiments of different knowledge-enhanced models.We found that leveraging windowed cross-attention and Mahalanobis-guided query construction to retrieve internal key knowledge from the LLM yields superior performance compared to directly employing a knowledge-augmented LLM for downstream tasks.
We guess the reason is that although knowledge augmented LLMs such as BioMixtral 7B encode medical knowledge through pretraining, they may not explicitly surface critical risk factors for specific knowledge. For instance, in the MIMIC-III mortality task, the model might miss the implication of structured features like \textit{mechanical ventilation} or \textit{high SOFA score} if not directly prompted. In contrast, our method retrieves relevant internal knowledge from the encoded medical graph, such as the relations between symptoms, interventions, and mortality and fuses it into the model input. This structured retrieval improves the model's ability to reason over clinical signals and enhances prediction accuracy.

\subsection{Comparison of Query Representation Strategies}
\label{con:query_re_stra}
\begin{table}[ht]

	\centering
	\renewcommand\arraystretch{1.3}
	\scalebox{0.6}{
	\begin{tabular} {c|c|cccc}
    \Xhline{0.5pt} 
		\hline 
       &Model &F1 &Jaccard& AUPRC &AUROC  \\
       \hline
  
             \multirow{3}{*}{Mortality-III}  & K2K   &  \textbf{18.55}   &\textbf{ 10.22 } &  \textbf{15.22  }& \textbf{61.05}   \\
                        &   K2K w Euclidean  &   16.97 &  9.27 &  9.67   &  57.25     \\
                           &  K2K (Mean Only)  &  12.06  &  6.42 &  8.45   &  52.51     \\
        \hline
         \multirow{3}{*}{Readmission-III}  & K2K   &   \textbf{ 69.31} &\textbf{ 53.03 } & \textbf{62.49}   & \textbf{ 56.64 } \\
                        &   K2K w Euclidean  & 63.27   & 46.28  &   58.26 &   53.25    \\
                           &  K2K (Mean Only) &  63.98  & 47.03  &  54.67   &  50.92     \\
        \hline
           \multirow{3}{*}{Mortality-IV}  & K2K   & \textbf{ 6.61}  & \textbf{3.42 }& \textbf{ 2.93  } & \textbf{66.50}    \\
                        &   K2K w Euclidean  & 4.79  & 2.45 &  2.19  &  61.81    \\
                           &  K2K (Mean Only) &  0.82  & 0.44  & 2.51    &  61.73    \\
                      \hline
            \multirow{3}{*}{Readmission-IV}  & K2K   & \textbf{63.75}   & \textbf{46.79}  &   \textbf{68.67 } & \textbf{ 66.47}   \\
                 &   K2K w Euclidean  &  63.56  &  46.59 &  67.87   &  66.41     \\
                           &   K2K (Mean Only)  & 56.26   & 39.14  &  67.71   &   65.58    \\
        \hline

	\end{tabular}
 }
		\caption{Comparison of K2K with different query construction methods.}
	\label{con:query}
\end{table}
Table~\ref{con:query} presents various query representation strategies for assessing the importance of each query vector within a window context. K2K (Euclidean) uses Euclidean distance for token weighting, whereas K2K (Mean Only) computes the window representation via simple mean pooling. Table~\ref{con:query} shows that our Mahalanobis-guided query representation consistently outperforms prior approaches. Unlike Euclidean distance, which treats all dimensions equally, our method accounts for per-dimension variance and emphasizes informative low-variance directions. This leads to more precise token weighting and better contextual representations. The results validate the effectiveness of variance-aware distance metrics in enhancing retrieval-informed reasoning.

\subsection{Comparison of Knowledge from Different LLM Layers}
\label{con:layers}

\begin{figure}[t]
        \centering
        \includegraphics[width=1\columnwidth]{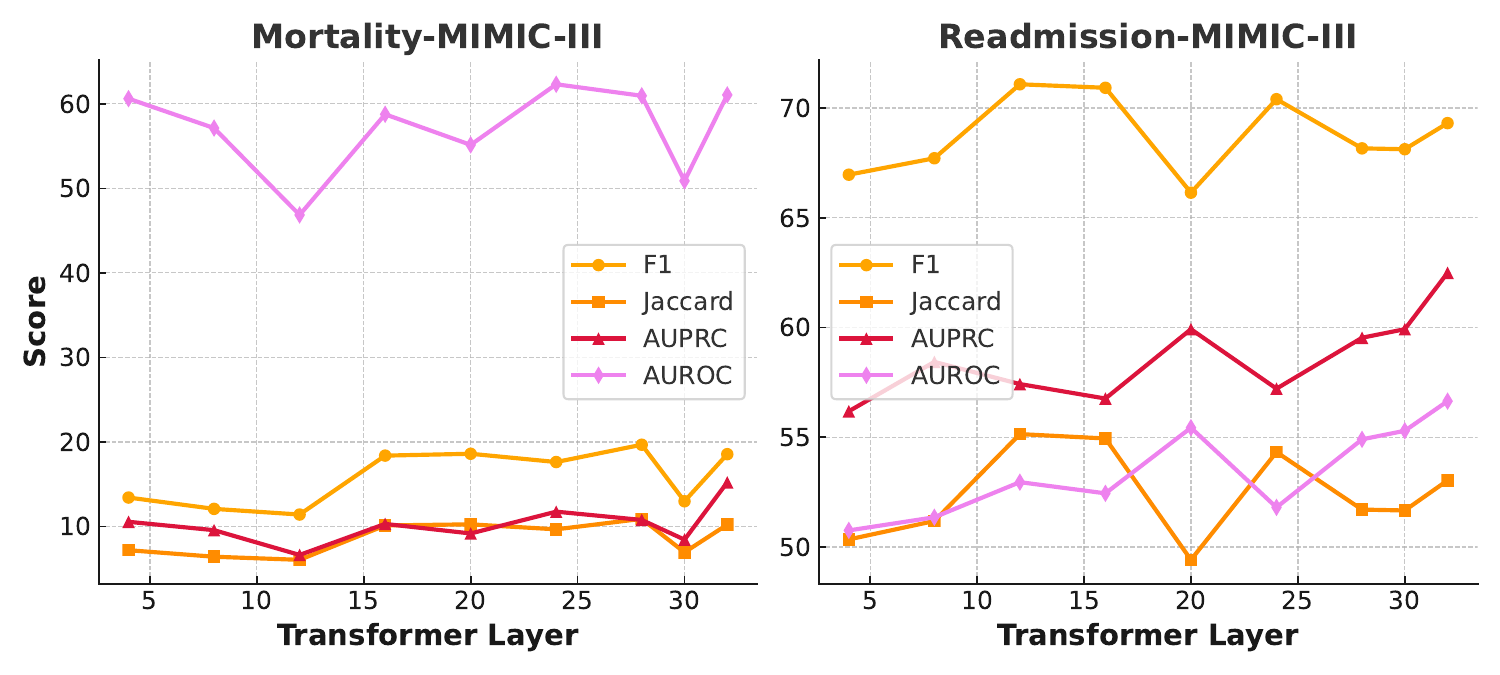}
	\caption{K2K performance with different layer knowledge. We used BioMistral-7B, which consists of 32 transformer layers. }
	\label{con:knowledge_source}
\end{figure}

In this section, we conduct experiments on K2K using knowledge (key) sources stored in different transformer layers within the LLM.  Both the document-based knowledge and the graph-based knowledge are extracted from the same corresponding layer. 
Figure~\ref{con:knowledge_source} reveals a nuanced deviation from the conventional view that upper layers in Transformers primarily encode semantic features while lower layers capture shallow, surface-level patterns. Although the final layers (e.g., Layer 30+) do contribute positively to performance in both Mortality-MIMIC-III and Readmission-MIMIC-III tasks, this improvement is not strictly monotonic. Notably, several shallow layers (e.g., Layers 5, 8, and 10) also exhibit strong performance across multiple metrics, indicating that valuable structural or entity-level knowledge resides in the lower layers as well. Furthermore, the impact of each layer varies across different evaluation metrics (F1, Jaccard, AUROC), suggesting that knowledge is distributed in a non-linear fashion throughout the network. These findings underscore the importance of considering both shallow and deep layers in knowledge extraction and reasoning tasks.

\section{Related Work}


Existing research has extensively explored leveraging retrieved information from diverse knowledge stores to enhance text understanding and generation~\cite{lewis2020retrieval,guu2020retrieval,li2023understand,li2025biomedrag,jiang2025reasoning}. These methods generally fall into two categories: structured and unstructured retrieval.

\textbf{Structured Knowledge Retrieval}: Several studies focus on integrating relational data. For instance, KIEST~\cite{li2023understand} dynamically injects entity and attribute knowledge from knowledge graphs to improve generation in entity-state-change tasks. Similarly, KARE~\cite{jiang2025reasoning} identifies relevant entities for each concept in a query and constructs a subgraph using shortest-path heuristics, providing structured relational context for downstream reasoning.

\textbf{Unstructured and Chunk-based Retrieval}: 
Other approaches utilize massive text corpora. REALM~\cite{guu2020retrieval} pioneered a gradient-based method to reward the retriever, thereby improving prediction accuracy through latent knowledge retrieval. More recently, BiomedRAG~\cite{li2025biomedrag} introduced a dynamic mechanism to rerank top-$k$ chunks from diverse biomedical databases. Similarly, RETRO~\cite{borgeaud2022improving} utilizes a chunk-based approach with cross-attention mechanisms to integrate retrieved segments from trillion-token scales.

Despite these advancements, traditional RAG remains burdened by the latency of querying massive, heterogeneous sources and the computational cost of processing lengthy external contexts. To mitigate these challenges, we propose a novel approach that retrieves knowledge directly from the internal key space of the LLM. By utilizing an activation-guided top-$k$ selection and a cross-window attention mechanism, our framework enables efficient, grounded, and low-latency knowledge access without the overhead of external search.

Recent work~\cite{xiao2024infllm, liu2024retrievalattention, fountas2025human} has explored retrieval modules that extract information from the key-value (KV) cache using probe queries derived from current context tokens. These methods typically treat the sliding window as a probe to retrieve relevant KV pairs. However, most overlook the critical role of probe construction, as LLMs are not inherently optimized for internal retrieval tasks.
To date, few studies have addressed how to optimize probe queries for internal key retrieval. One notable exception is ActQKV~\cite{xiao2025activation}, which proposes an activation-aware mechanism that selects key tokens based on activation magnitude and employs Euclidean distance for retrieval. However, this approach assumes equal importance across all embedding dimensions, thereby ignoring per-dimension variance and reducing sensitivity to meaningful deviations in low-variance directions. 

\section{Conclusion}

In this paper, we introduce K2K, a framework that retrieves knowledge directly from an LLM’s internal key space, bypassing context-heavy prompting and retrieval-intensive pipelines used in conventional RAG. With Mahalanobis-guided query representations and cross-attentive reranking for multi-source integration, K2K improves reasoning in knowledge-intensive clinical tasks. Our results show that LLM parameters are not merely latent carriers of knowledge but can be explicitly accessed to enhance predictive performance.
\section{Limitations}

While the K2K framework demonstrates robust performance in internal knowledge retrieval and integration, several limitations remain to be addressed in future work.

\textbf{Layer Selection and Granularity}: The retrieval memory is currently constructed from fixed layers of the pre-trained model. Although LoRA-based infusion facilitates domain adaptation, our approach does not yet dynamically select which specific layers or representations (e.g., lower-level semantic features vs. higher-level abstraction in FFN layers) are most informative for a given query. Implementing a layer-wise selection or multi-layer weighting mechanism could further improve retrieval fidelity and computational efficiency.

\textbf{Domain Generalization}: Our framework has been primarily evaluated within the biomedical domain. While this provides a rigorous testbed for high-stakes decision-making, the generalizability of K2K to other knowledge-intensive fields—such as legal reasoning or financial forecasting—remains to be explored. Furthermore, while K2K performs well on benchmark datasets, future iterations must address the pervasive issue of data imbalance inherent in real-world clinical tasks to ensure equitable performance across rare medical conditions.

\bibliography{custom}

\appendix


\section{Appendices}
\subsection{Effect of LLM Backbone Choice}
\label{Effect_of_LLM_Choice}

\subsection{Implementation Detail}
\label{con:Implementation Detail}

The chunk size is set to 64 throughout this work. For the top-$k$ values, we use $k=5$ for Mortality-MIMIC-III, $k=20$ for Readmission-MIMIC-III and Mortality-MIMIC-IV, and $k=10$ for Readmission-MIMIC-IV.  The same LLM backbone is used during both the retrieval phase, when keys are extracted, and the training/inference phases, when those keys are utilized, ensuring alignment in the representation space.    We use AdamW as our optimizer, with a learning rate of $2 \times 10^{-5}$ and $\epsilon$ set to $1 \times 10^{-8}$. The batch size is 16.
For the cross-attention module, we set the model dimension to 4096 and apply a dropout rate of 0.3.

\subsection{Separately retrieval}
We intentionally use only the base component $W_1$ from the final FFN layer of $\mathcal{M}_{\text{domain}}^{\text{doc}}$ to represent document knowledge. This design is motivated by the need to preserve a clear and interpretable separation between knowledge sources. Specifically, (1) \textbf{theoretically}, unstructured document knowledge (captured by $W_1$) and structured graph knowledge (injected via $AB$) differ fundamentally in format and reasoning mechanisms, and thus should not be merged directly in representation; (2) \textbf{in practice}, combining them into a single matrix $W_1 + AB$ would entangle their contributions, making it difficult to analyze or attribute model behavior to specific knowledge types; and (3) \textbf{from an engineering perspective}, separating the two enables more modular system design, facilitates ablation studies, debugging, incremental updates, and future knowledge extension.

\subsection{Mahalanobis distance}
\subsection*{Step 1: Compute the Covariance Matrix \( \Sigma \)}
$$
\Sigma = \frac{1}{L - 1} \sum_{j=1}^{L} (q_j^w - \bar{z}^w)(q_j^w - \bar{z}^w)^T \quad \in \mathbb{R}^{D \times D}
$$

\subsection*{Step 2: Compute the Mahalanobis Distance (Activation Bias) \( \phi_j^t \)}
$$
\phi_j^w= \sqrt{(q_j^w - \bar{z}^w)^T \Sigma^{-1} (q_j^w - \bar{z}^w)} \quad \in \mathbb{R}
$$

\subsection*{Step 3: Construct the Probe-Query Vector \( \mathbf{Q}_{\text{probe}}^w\)}
$$
\mathbf{Q}_{\text{probe}}^w= \sum_{j=1}^{L} \alpha_j^w \cdot q_j^w, 
\quad \text{where} \quad 
\alpha_j^w = \frac{\phi_j^w}{\sum_{j=1}^{L} \phi_j^w} 
$$

\subsection{Comparison of K2K with different chunk size}
\label{con:chunk_size_compare}

\begin{figure}[t]
        \centering
        \includegraphics[width=1\columnwidth]{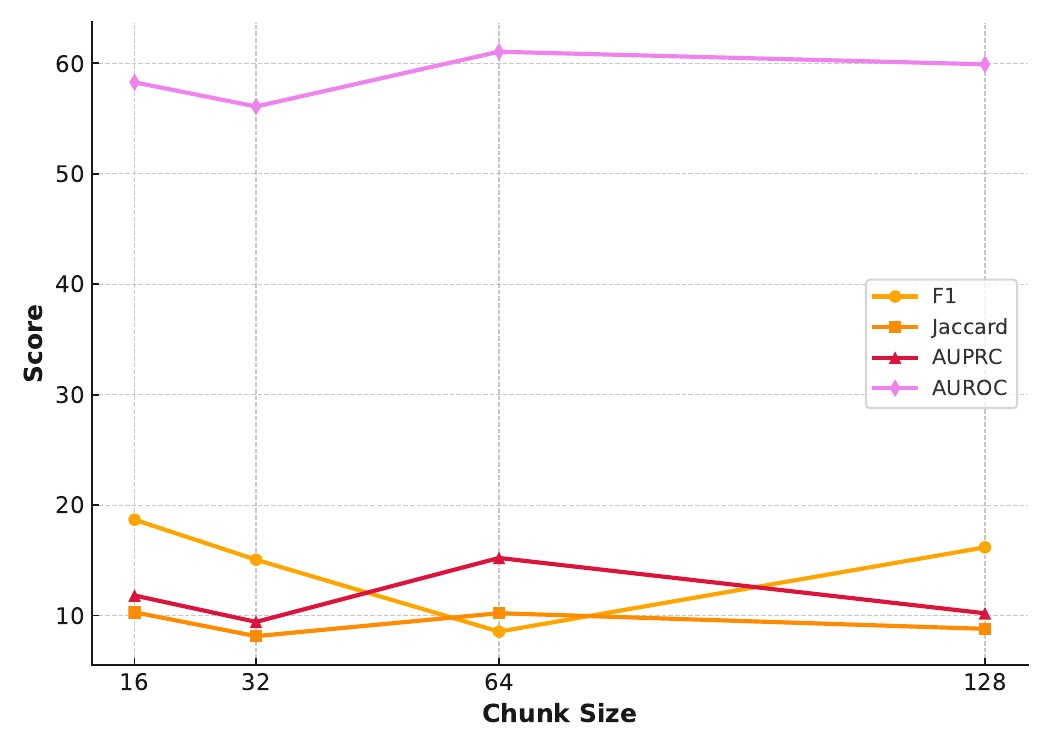}
	\caption{K2K performance with different chunk sizes. }
	\label{con:chunk_size}
\end{figure}

Figure~\ref{con:chunk_size} shows the K2K performance of different chunk sizes on the dataset MIMIC-III Mortality. We choose four chunk sizes: 16, 32, 64, and 128. We observe that smaller chunk sizes (e.g., 16) lead to higher F1 scores, indicating that finer granularity benefits the identification of relevant knowledge segments. However, chunk size 64 achieves the highest AUPRC and AUROC, suggesting it better balances precision and recall for more robust classification. Larger chunk sizes may reduce retrieval frequency but risk diluting critical signals. Therefore, chunk size selection should consider both task sensitivity and retrieval efficiency.

\subsection{Ablation Study on Top-$k$ Retrieval}
\label{con:topk_ana}
\begin{figure}[t]
        \centering
        \includegraphics[width=1\columnwidth]{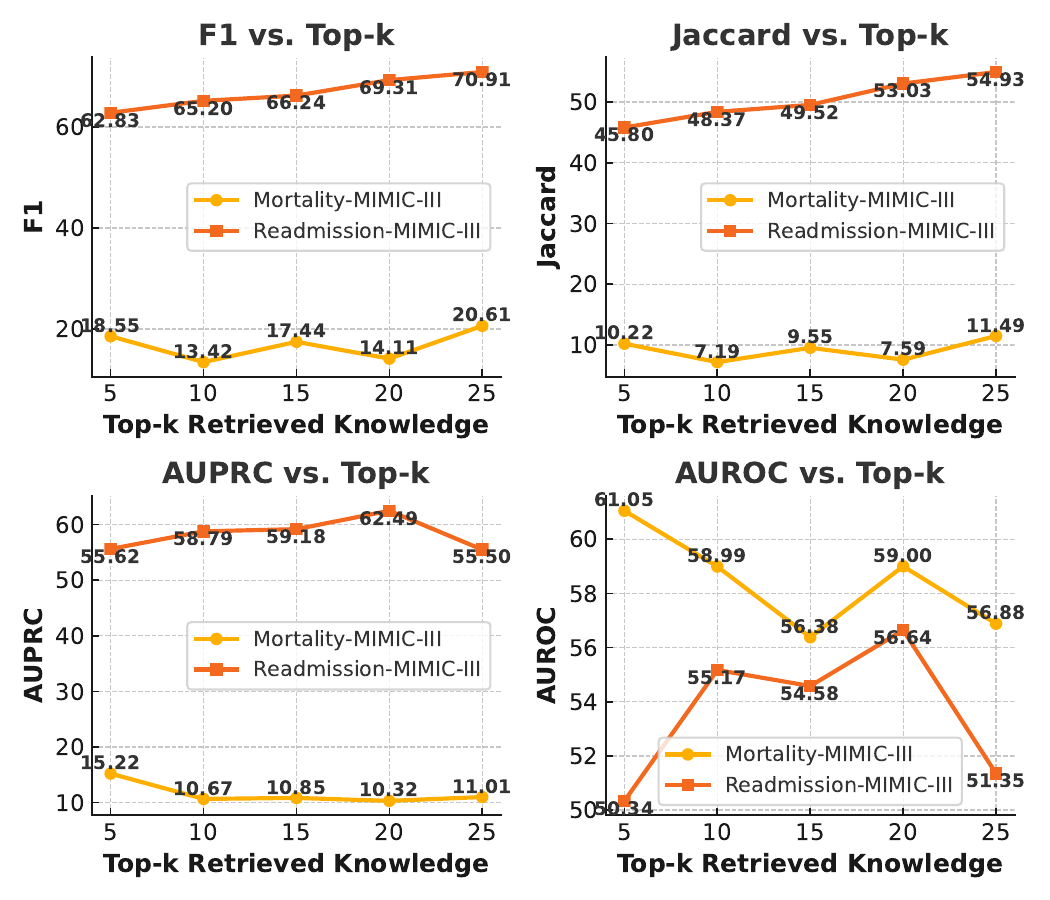}
	\caption{K2K Performance Across Different Top-k Retrieved Knowledge Values on MIMIC-III.}
	\label{con:mimiciii_topk}
\end{figure}

\begin{figure}[t]
        \centering
        \includegraphics[width=1\columnwidth]{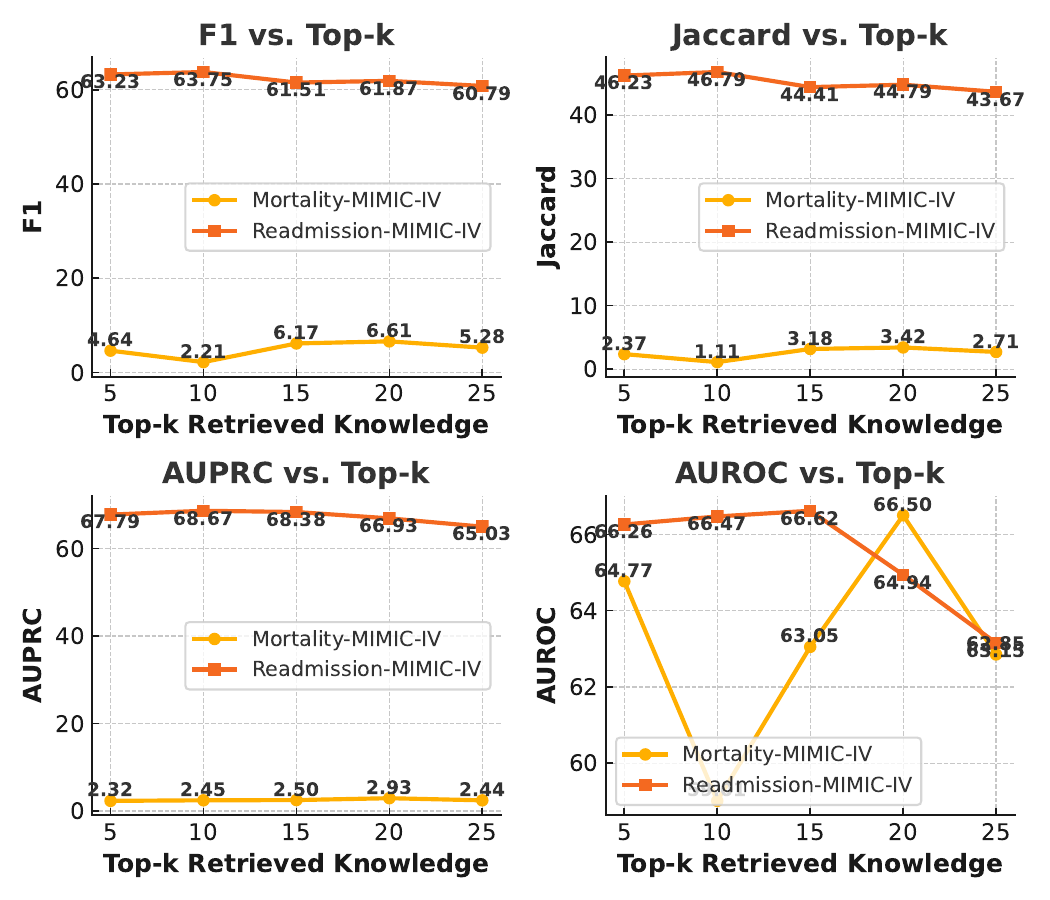}
	\caption{K2K Performance Across Different Top-k Retrieved Knowledge Values on MIMIC-IV. }
	\label{con:mimiciv_topk}
\end{figure}

Figures~\ref{con:mimiciii_topk} and ~\ref{con:mimiciv_topk} demonstrate how the number of retrieved knowledge entries (top-k) affects the performance of K2K on both MIMIC-III and MIMIC-IV datasets. For MIMIC-III, performance generally improves with increasing top-k, with the best F1 (20.61) and Jaccard (11.49) observed at k=25 for the mortality task, while the readmission task achieves optimal results at k=20–25. Notably, AUROC and AUPRC peak at k=20, suggesting a balance between sufficient context and noise control.

In contrast, for MIMIC-IV, mortality prediction shows a performance peak at k=20 across all metrics, particularly for F1 and AUPRC, while readmission results are relatively stable across k values, with the highest F1 (63.75) and AUPRC (68.67) at k=10. However, large k values (e.g., k=25) tend to hurt AUROC, especially for readmission. These results indicate that task-specific tuning of top-k is crucial, and that mortality prediction benefits more from increasing top-k, while readmission may require a smaller, more focused knowledge set.

\subsection{Retrieval Efficiency Comparison Across different Retrieval methods}
\label{con:Retrieval_Inference_Efficiency}
\begin{table}[t]
\centering
\renewcommand{\arraystretch}{1.15}
\setlength{\tabcolsep}{6pt}
\scalebox{0.80}{
\begin{tabular}{l cc}
\Xhline{0.6pt}

\textbf{}
& \textbf{Avg}
& \textbf{Retrieval Time} \\

\Xhline{0.4pt}

KARE & 21.11 & 00:33:52   \\


Prompt-based
& 22.52 & 3:26:00   \\

K2K
& 22.89 & 0:0:5\\

\Xhline{0.6pt}
\end{tabular}
}

\caption{Performance comparison on the MIMIC-III dataset in terms of averaged metrics (Avg), retrieval time.}
\label{tab:mimic3_efficiency}
\end{table}

Compared to prior retrieval approaches, our K2K method demonstrates substantially higher efficiency. Specifically, KARE performs multi-stage reasoning by first retrieving co-existing concepts appear in each patient’s data and then computing the shortest paths between the concepts and the co-existing concepts over a large knowledge graph. This results in a total complexity of $O(k(|V| + |E|))$, where $k$ is the number of co-existing concepts, and $|V|, |E|$ are the number of nodes and edges in the graph, respectively. Contriever, a dense retriever, encodes the query and computes similarities across the entire corpus, resulting in a time complexity of $O(Nd)$ without approximation, where $N$ is the number of documents and $d$ is the embedding dimension. Prompt-based retrieval avoids external indexing but relies on LLM generation conditioned on carefully designed instructions, which incurs substantial inference cost at $O(L \cdot n^2 \cdot h)$, where $L$ is the number of layers, $n$ is the token length, and $h$ is the number of attention heads.

In contrast, K2K bypasses both external document retrieval and graph traversal by directly reusing the internal knowledge of the LLM. It retrieves relevant knowledge by comparing current input representations with pre-trained FFN keys and LoRA adapter keys from a specific transformer layer. This enables fast memory access with a time complexity of only $O(m)$ or $O(mk)$ (for top-k selection), where $m$ is the number of tokens. By removing the need for external retrieval or generation, K2K achieves the fastest inference speed among all retrievers while maintaining high accuracy, demonstrating the efficiency and practicality of internal knowledge utilization. 

\end{document}